%% file: main.tex
\documentclass[10pt,twocolumn,letterpaper]{article}

\usepackage[pagenumbers]{cvpr} 

\input{preamble}

\usepackage{graphicx}
\usepackage{float}

\usepackage{array}
\usepackage{lipsum} 
\usepackage{graphicx}
\usepackage{enumitem}
\usepackage{subcaption}

\usepackage{amsthm}

\usepackage{mathtools}
\usepackage{ragged2e}
\usepackage{xpatch}
\usepackage{soul}
\usepackage[normalem]{ulem}
\usepackage{makecell}

\usepackage{colortbl}
\usepackage[ruled, linesnumbered, noend]{algorithm2e}
\usepackage{algpseudocode}%

\usepackage{pifont}
\newcommand{\cmark}{\ding{51}}%
\newcommand{\xmark}{\ding{55}}%

\definecolor{cvprblue}{rgb}{0.21,0.49,0.74}
\usepackage[pagebackref,breaklinks,colorlinks,citecolor=cvprblue]{hyperref}

\def\confName{CVPR}
\def\confYear{2024}

\title{DeepCache: Accelerating Diffusion Models for Free}

\author{
\bf Xinyin Ma \quad Gongfan Fang \quad Xinchao Wang\thanks{Corresponding author}\\
{National University of Singapore} \\
{\tt\small \{maxinyin, gongfan\}@u.nus.edu, xinchao@nus.edu.sg} \\
}

\newcommand{\methodname}{DeepCache}

\newcommand\blfootnote[1]{%
  \begingroup
  \renewcommand\thefootnote{}\footnote{#1}%
  \addtocounter{footnote}{-1}%
  \endgroup
}

\begin{document}

\twocolumn[{%
\renewcommand\twocolumn[1][]{#1}%
\maketitle
\begin{center}
    \centering
    \captionsetup{type=figure}
    \vspace{-27pt}
    \includegraphics[width=\textwidth]{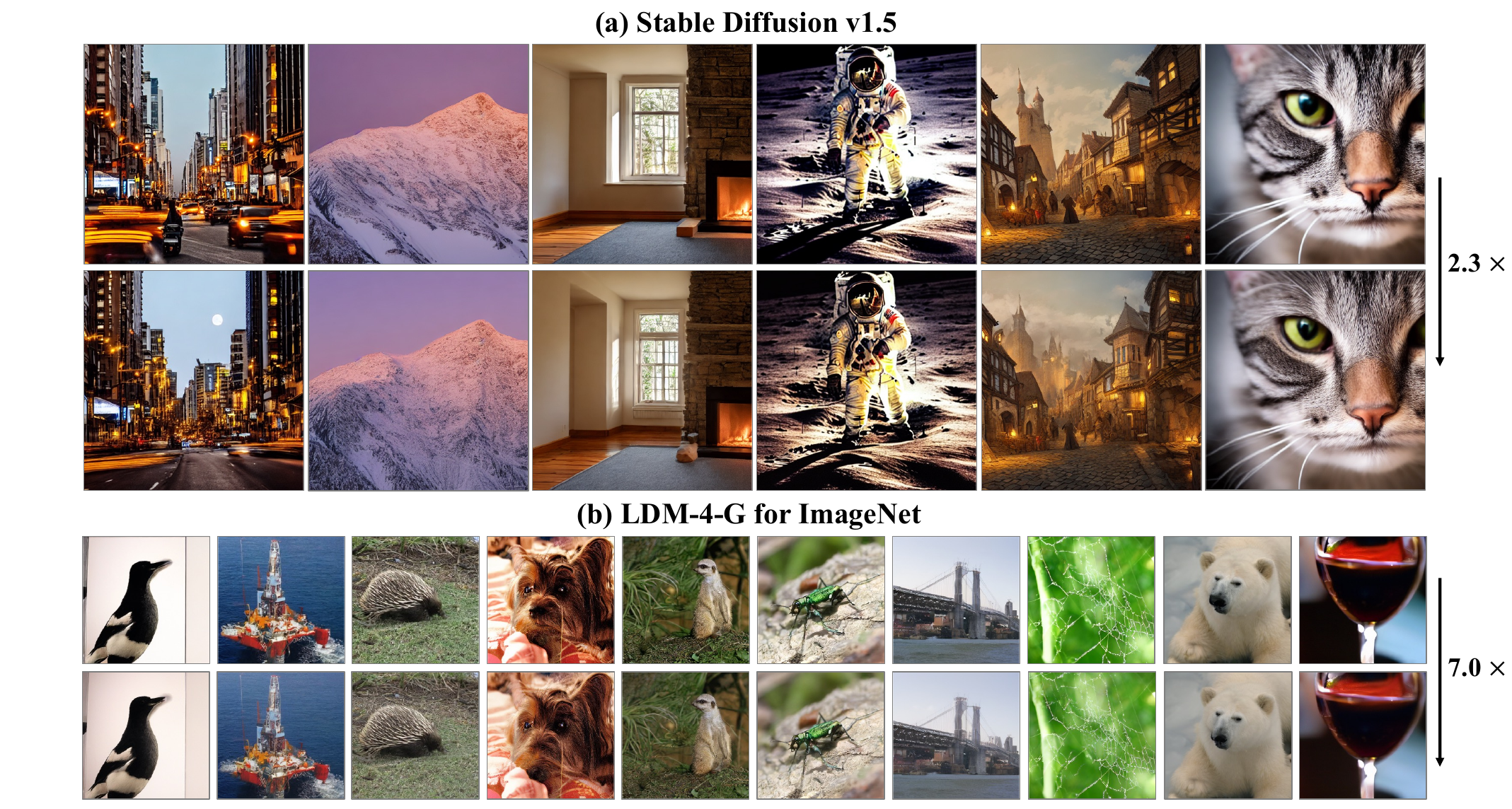}
    \vspace{-18pt}
    \caption{
    \label{fig:teaser}
    Accelerating Stable Diffusion V1.5 and LDM-4-G by 2.3$\times$ and 7.0$\times$, with 50 PLMS steps and 250 DDIM steps respectively.} 
    
\end{center}%
}]

\maketitle

\input{sec/0_abstract} 
\input{sec/1_intro.tex}

\input{sec/2_relatedwork.tex}

\input{sec/3_method.tex}

\input{sec/experiment}

\clearpage

{
    \small
    \bibliographystyle{ieeenat_fullname}
    \bibliography{main}
}


\input{sec/X_suppl}

\end{document}

%% file: preamble.tex
\usepackage[dvipsnames]{xcolor}


%% file: sec/0_abstract.tex
\begin{abstract}
Diffusion models have recently gained unprecedented attention in the field of image synthesis due to their remarkable generative capabilities. Notwithstanding their prowess, these models often incur substantial computational costs, primarily attributed to the sequential denoising process and cumbersome model size. Traditional methods for compressing diffusion models typically involve extensive retraining, presenting cost and feasibility challenges. In this paper, we introduce DeepCache, a novel training-free paradigm that accelerates diffusion models from the perspective of model architecture. DeepCache capitalizes on the inherent temporal redundancy observed in the sequential denoising steps of diffusion models, which caches and retrieves features across adjacent denoising stages, thereby curtailing redundant computations. Utilizing the property of the U-Net, we reuse the high-level features while updating the low-level features in a very cheap way. This innovative strategy, in turn, enables a speedup factor of 2.3$\times$ for Stable Diffusion v1.5 with only a 0.05 decline in CLIP Score, and 4.1$\times$  for LDM-4-G with a slight decrease of 0.22 in FID  on ImageNet. Our experiments also demonstrate DeepCache's superiority over existing pruning and distillation methods that necessitate retraining and its compatibility with current sampling techniques. Furthermore, we find that under the same throughput, DeepCache effectively achieves comparable or even marginally improved results with DDIM or PLMS. Code is available at \url{https://github.com/horseee/DeepCache}. \blfootnote{$^*$ Corresponding author}

\end{abstract}

%% file: sec/1_intro.tex
\section{Introduction}
\label{sec:intro}
In recent years, diffusion models \cite{sohl2015deep,song2019generative,ho2020ddpm,dhariwal2021adm} have emerged as a pivotal advancement in the field of generative modeling, gaining substantial attention for their impressive capabilities. These models have demonstrated remarkable efficacy across diverse applications, being employed for the generation of images \cite{song2020sliced,vahdat2021score,kawar2022denoising}, text \cite{li2022diffusion,gong2022diffuseq}, audio \cite{chen2020wavegrad,popov2021gradtts}, and video \cite{singer2022make,ho2022imagen,luo2023videofusion}. A large number of attractive applications have been facilitated with diffusion models, including but not limited to image editing~\cite{avrahami2022blended,kawar2023imagic,meng2021sdedit}, image super-enhancing~\cite{saharia2022image,li2022srdiff}, image-to-image translation~\cite{choi2021ilvr,saharia2022palette}, text-to-image generation~\cite{nichol2021glide,saharia2022photorealistic,rombach2022high,ramesh2022hierarchical} and text-to-3D generation~\cite{luo2021diffusion,lin2023magic3d,poole2022dreamfusion}. 

Despite the significant effectiveness of diffusion models, their relatively slow inference speed remains a major obstacle to broader adoption, as highlighted in~\cite{li2023snapfusion}. The core challenge stems from the step-by-step denoising process required during their reverse phase, limiting parallel decoding capabilities~\cite{shih2023parallel}. Efforts to accelerate these models have focused on two main strategies: reducing the number of sampling steps, as seen in approaches ~\cite{song2020ddim,lu2022dpmsolver,salimans2022progressive,meng2023distillation}, and decreasing the model inference overhead per step through methods like model pruning, distillation, and quantization~\cite{fang2023structural,kim2023bksdm,he2023ptqd}.

Our goal is to enhance the efficiency of diffusion models by reducing model size at each step. 
Previous compression methods for diffusion models focused on re-designing network architectures through a comprehensive structural analysis~\cite{li2023snapfusion} or involving frequency priors into the model design~\cite{yang2023diffusion}, which yields promising results on image generation. However, they require large-scale datasets for retraining these lightweight models. Pruning-based methods, as explored by ~\cite{fang2023structural,kim2023bksdm}, lessen the data and training requirements to 0.1\% of the full training. Alternatively, ~\cite{liu2023oms} employs adaptive models for different steps, which is also a potential solution. However, it depends on a collection of pre-trained models or requires optimization of sub-networks~\cite{yang2023ddsm}. Those methods can reduce the expense of crafting a new lightweight model, but the retraining process is still inevitable, which makes the compression costly and less practical for large-scale pre-trained diffusion models, such as Stable Diffusion~\cite{rombach2022ldm}. 

To this end, we focus on a challenging topic: \emph{How to significantly reduce the computational overhead at each denoising step without additional training, thereby achieving a cost-free compression of Diffusion Models?} 
To achieve this, we turn our attention to the intrinsic characteristics of the reverse denoising process of diffusion models, observing a significant temporal consistency in the high-level features between consecutive steps. We found that those high-level features are even cacheable, which can be calculated once and retrieved again for the subsequent steps. By leveraging the structural property of U-Net, the high-level features can be cached while maintaining the low-level features updated at each denoising step.
Through this, a considerable enhancement in the efficiency and speed of Diffusion Models can be achieved without any training.

To summarize, we introduce a novel paradigm for the acceleration of Diffusion Models, which gives a new perspective for training-free accelerating the diffusion models. It is not merely compatible with existing fast samplers but also shows potential for comparable or superior generation capabilities. The contributions of our paper include:
\begin{itemize}
    \item We introduce a simple and effective acceleration algorithm, named \methodname, for dynamically compressing diffusion models during runtime and thus enhancing image generation speed without additional training burdens.
    \item \methodname\ utilizes the temporal consistency between high-level features. With the cacheable features, the redundant calculations are effectively reduced. Furthermore, we introduce a non-uniform 1:N strategy, specifically tailored for long caching intervals.
    \item \methodname\ is validated across a variety of datasets, including CIFAR, LSUN-Bedroom/Churches, ImageNet, COCO2017 and PartiPrompt, and tested under  DDPM, LDM, and Stable Diffusion. Experiments demonstrate that our approach has superior efficacy than pruning and distillation algorithms that require retraining under the same throughput.
\end{itemize}

%% file: sec/2_relatedwork.tex
\section{Related Work}

\label{sec:related_work}


High-dimensional image generation has evolved significantly in generative modeling. Initially, GANs \cite{goodfellow2014gan,arjovsky2017wassersteingan} and VAEs \cite{kingma2013vae,higgins2016betavae} led this field but faced scalability issues due to instability and mode collapse \cite{kodali2017ganstability}. Recent advancements have been led by Diffusion Probabilistic Models \cite{ho2020ddpm,dhariwal2021adm,rombach2022ldm,song2020score}, which offer superior image generation. However, the inherent nature of the reverse diffusion process \cite{song2019generative} slows down inference. Current research is focused on two main methods to speed up diffusion model inference.

\paragraph{Optimized Sampling Efficiency.} focuses on reducing the number of sampling steps. 
DDIM ~\cite{song2020ddim} reduces these steps by exploring a non-Markovian process, related to neural ODEs. Studies ~\cite{lu2022dpmsolver,zhang2022fast,bao2022analytic,liu2022pseudo} further dive into the fast solver of SDE or ODE to create efficient sampling of diffusion models. Some methods progressively distilled the model to reduced timestep~\cite{salimans2022progressive} or replace the remaining steps with a single-step VAE~\cite{lyu2022accelerating}. The Consistency Model~\cite{song2023consistency} converts random noise to the initial images with only one model evaluation. Parallel sampling techniques like DSNO ~\cite{zheng2023dsno} and ParaDiGMS ~\cite{shih2023parallel} employ Fourier neural operators and Picard iterations for parallel decoding
.

\paragraph{Optimized Structural Efficiency.} This approach aims to reduce inference time at each sampling step. It leverages strategies like structural pruning in Diff-pruning~\cite{fang2023structural} and efficient structure evolving in SnapFusion~\cite{li2023snapfusion}. Spectral Diffusion~\cite{yang2023diffusion} enhances architectural design by incorporating frequency dynamics and priors. In contrast to these methods, which use a uniform model at each step, \cite{liu2023oms} proposes utilizing different models at various steps, selected from a diffusion model zoo. The early stopping mechanism in diffusion is explored in~\cite{li2023autodiffusion,moon2023early,tang2023deediff}, while the quantization techniques~\cite{he2023ptqd,shang2023post} focus on low-precision data types for model weights and activations. Additionally, \cite{batzolis2022non} and \cite{bolya2023token} present novel approaches to concentrate on inputs, with the former adopting a unique forward process for each pixel and the latter merging tokens based on similarity to enhance computational efficiency in attention modules.
Our method is categorized under an objective to minimize the average inference time per step. Uniquely, our approach reduces the average model size substantially for each step, accelerating the denoising process without necessitating retraining.

%% file: sec/3_method.tex
\section{Methodology}
\subsection{Preliminary}

\paragraph{Forward and Reverse Process.} Diffusion models \cite{ho2020ddpm} simulate an image generation process using a series of random diffusion steps. The core idea behind diffusion models is to start from random noise and gradually refine it until it resembles a sample from the target distribution. 
In the forward diffusion process, with a data point sampled from the real distribution, $\mathbf{x}_0 \sim q(\mathbf{x})$, gaussian noises are gradually added in T steps:
\begin{align}
    q\left(\mathbf{x}_t | \mathbf{x}_{t-1}\right)=\mathcal{N}\left(\mathbf{x}_t ; \sqrt{1-\beta_t} \mathbf{x}_{t-1}, \beta_t \mathbf{I}\right)
\end{align}
where $t$ is the current timestep and $\{\beta_t\}
$ schedules the noise. The \textit{reverse diffusion process} denoises the random noise $x_T \sim \mathcal{N}(0, \mathbf{I})$ into the target distribution by modeling $q\left(\mathbf{x}_{t-1} | \mathbf{x}_{t}\right)$. At each reverse step $t$, the conditional probability distribution is approximated by a network $\epsilon_\theta\left(\mathbf{x}_t, t\right)$ with the timestep $t$ and previous output $\mathbf{x}_t$ as input:
\begin{align}
    & x_{t-1} \sim p_{\theta}(x_{t-1}|x_t) = \nonumber \\
    & \mathcal{N}\left(x_{t-1}; \frac{1}{\sqrt{\alpha_t}}\left(\mathbf{x}_t-\frac{1-\alpha_t}{\sqrt{1-\bar{\alpha}_t}} \boldsymbol{\epsilon}_\theta\left(\mathbf{x}_t, t\right)\right), \beta_t\mathbf{I}\right) 
\end{align}
where $\alpha_t = 1 - \beta_t$ and $\bar{\alpha}_t = \prod_{i=1}^T \alpha_i$. Applied iteratively, it gradually reduces the noise of the current $\mathbf{x}_t$, bringing it close to a real data point when we reach $x_0$. 

\paragraph{High-level and Low-level Features in U-Net.} 

U-Net~\cite{ronneberger2015unet} was originally introduced for biomedical image segmentation and showcased a strong ability to amalgamate low-level and high-level features, attributed to the skip connections. U-Net is constructed on stacked downsampling and upsampling blocks, which encode the input image into a high-level representation and then decode it for downstream tasks. The block pairs, denoted as $\{D_i\}_{i=1}^{d}$ and $\{U_i\}_{i=1}^{d}$, are interconnected with additional skip paths. Those skip paths directly forward the rich and relatively more low-level information from $D_i$ to $U_i$. During the forward propagation in the U-Net architecture, the data traverses concurrently through two pathways: the \textit{main branch} and the \textit{skip branch}. These branches converge at a concatenation module, with the \textit{main branch} providing processed high-level feature from the preceding upsampling block $U_{i+1}$, and the \textit{skip branch} contributing corresponding feature from the symmetrical block $D_i$. Therefore, at the heart of a U-Net model is a concatenation of low-level features from the skip branch, and the high-level features from the main branch, formalized as:
\begin{equation}
    \operatorname{Concat}(D_i(\cdot), U_{i+1}(\cdot))
\end{equation}

\begin{figure*}[t]
    \centering
    \resizebox{1.0\linewidth}{!}{
    \includegraphics{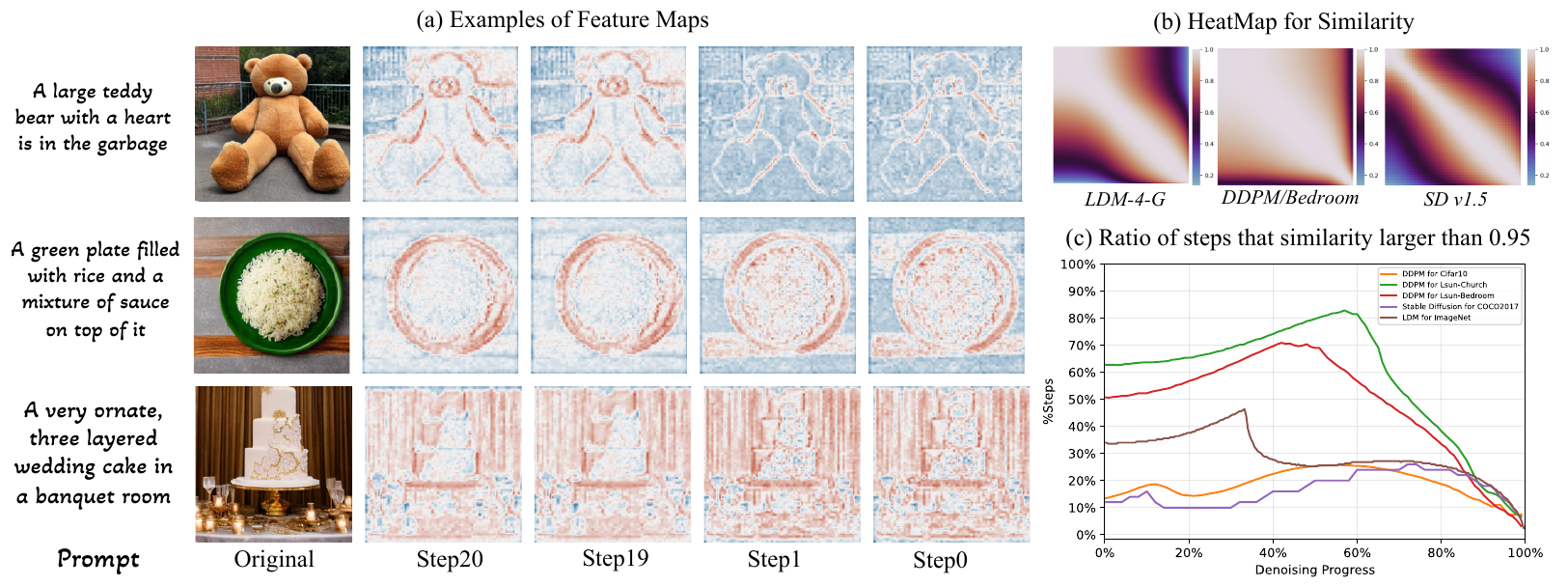}
    }
    \caption{(a) Examples of feature maps in the up-sampling block $U_2$ in Stable Diffusion. We present a comparison from two adjacently paired steps, emphasizing the invariance inherent in the denoising process. (b) Heatmap of similarity between $U_2$'s features in all steps on three typical diffusion models. (c) The percentage of steps with a similarity greater than 0.95 to the current step.}
    \label{fig:similarity}
    \vspace{-2mm}
\end{figure*}

\subsection{Feature Redundancy in Sequential Denoising}
The inherent sequentiality of the denoising process in diffusion models presents a primary bottleneck in inference speed. Previous methods primarily employed strategies that involved skipping certain steps to address this issue. In this work,  we revisit the entire denoising process, seeking to uncover specific properties that could be optimized to enhance inference efficiency. 

\newtheorem*{definition}{Observation}

\begin{definition}[]
Adjacent steps in the denoising process exhibit significant temporal similarity in high-level features.
\end{definition}
In Figure \ref{fig:similarity}, we provide empirical evidence related to this observation. The experiments elucidate two primary insights: 1) There is a noticeable temporal feature similarity between adjacent steps within the denoising process, indicating that the change between consecutive steps is typically minor; 2) Regardless of the diffusion model we used, for each timestep, at least 10\% of the adjacent timesteps exhibit a high similarity ($>$0.95) to the current step, suggesting that certain high-level features change at a gradual pace. This phenomenon can be observed in a large number of well-established models like Stable Diffusion, LDM, and DDPM. In the case of DDPM for LSUN-Churches and LSUN-Bedroom, some timesteps even demonstrate a high degree of similarity to 80\% of the other steps, as highlighted in the green line in Figure \ref{fig:similarity} (c).

Building upon these observations, our objective is to leverage this advantageous characteristic to accelerate the denoising process. Our analysis reveals that the computation often results in a feature remarkably similar to that of the previous step, thereby highlighting the existence of redundant calculations for optimization. We contend that allocating significant computational resources to regenerate these analogous feature maps constitutes an inefficiency.  While incurring substantial computational expense, yields marginal benefits, it suggests a potential area for efficiency improvements in the speed of diffusion models.

\subsection{Deep Cache For Diffusion Models}

We introduce \methodname, a simple and effective approach that leverages the temporal redundancy between steps in the reverse diffusion process to accelerate inference. Our method draws inspiration from the caching mechanism in a computer system, incorporating a storage component designed for elements that exhibit minimal changes over time. Applying this in diffusion models, we eliminate redundant computations by strategically caching slowly evolving features, thereby obviating the need for repetitive recalculations in subsequent steps.


To achieve this, we shift our focus to the skip connections within U-Net, which inherently offers a dual-pathway advantage: 
the main branch requires heavy computation to traverse the entire network, while the skip branch only needs to go through some shallow layers, resulting in a very small computational load.
The prominent feature similarity in the main branch, allows us to reuse the already computed results rather than calculate it repeatedly for all timesteps.



\begin{figure}[t]
    \centering
    \includegraphics[width=\linewidth]{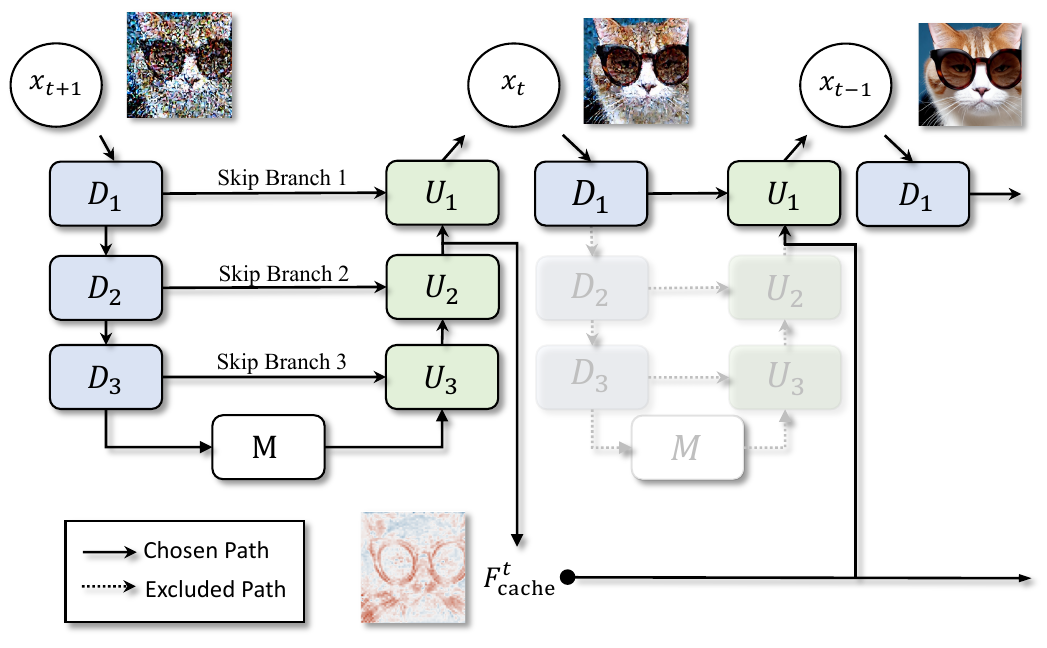}
    \caption{Illustration of DeepCache. At the $t-1$ step, $x_{t-1}$ is generated by reusing the features cached at the $t$ step, and the blocks ${D_2, D_3, U_2, U_3}$ are not executed for more efficient inference.}
    \label{fig:method}
\end{figure}

\paragraph{Cacheable Features in denosing.}
To make this idea more concrete, we study the case within two consecutive timesteps $t$ and $t-1$. According to the reverse process, $x_{t-1}$ would be conditionally generated based on the previous results $x_{t}$. First, we generate $x_t$ in the same way as usual, where the calculations are performed across the entire U-Net. To obtain the next output $x_{t-1}$, we retrieve the high-level features produced in the previous $x_t$. More specifically, consider a skip branch $m$ in the U-Net, which bridges $D_m$ and $U_{m}$, we cache the feature maps from the previous up-sampling block at the time $t$ as the following:
\begin{align}
     F^t_{\text{cache}} \leftarrow U_{m+1}^t(\cdot) \label{eqn:fearture_cache}
\end{align}
which is the feature from the main branch at timestep $t$. Those cached features will be plugged into the network inference in the subsequent steps. In the next timestep $t-1$, the inference is not carried out on the entire network; instead, we perform a dynamic partial inference. Based on the previously generated $x_t$, we only calculate those that are necessary for the $m$-th skip branch and substitute the compute of the main branch with a retrieving operation from the cache in Equation \ref{eqn:fearture_cache}. Therefore, the input for $U^{t-1}_m$ in the $t-1$ timestep can be formulated as:
\begin{align}
    \operatorname{Concat}(D^{t-1}_m(\cdot), F_{\text{cache}}^t)
\end{align}
Here, $D^{t-1}_m$ represents the output of the $m$-th down-sampling block, which only contains a few layers if a small $m$ is selected. For example, if we perform DeepCache at the first layer with $m=1$, then we only need to execute one downsampling block to obtain $D^{t-1}_1$. As for the second feature $F_{\text{cache}}^t$, no additional computational cost is needed since it can be simply retrieved from the cache. We illustrate the above process in Figure \ref{fig:method}. 

\paragraph{Extending to 1:N Inference} This process is not limited to the type with one step of full inference followed by one step of partial inference. As shown in Figure \ref{fig:similarity}(b), pair-wise similarity remains a high value in several consecutive steps. The mechanism can be extended to cover more steps, with the cached features calculated once and reused in the consecutive $N-1$ steps to replace the original $U^{t-n}_{m+1}(\cdot)$, $n \in \{1,\ldots,N-1\}$.
Thus, for all the T steps for denoising, the sequence of timesteps that performs full inference are: 
\begin{align}
    \mathcal{I} = \left\{ x \in \mathbb{N} \,|\, x = iN, \, 0 \leq i < k \right\}
\end{align}
where $k = \lceil T / N \rceil$ denotes the times for cache updating. 


\paragraph{Non-uniform 1:N Inference}
Based on the 1:N strategy, we managed to accelerate the inference of diffusion under a strong assumption that the high-level features are invariant in the consecutive N step. However, it's not invariably the case, especially for a large N, as demonstrated by
the experimental evidence in Figure \ref{fig:similarity}(c). The similarity of the features does not remain constant across all steps. For models such as LDM, the temporal similarity of features would significantly decrease around 40\% of the denoising process. Thus, for the non-uniform 1:N inference, we tend to sample more on those steps with relatively small similarities to the adjacent steps. Here, the sequence of timesteps to perform full inference becomes: 
\begin{align}
    \mathcal{L} &= \left\{l_i \mid l_i \in \operatorname{linear\_space}\left((-c)^{\frac{1}{p}}, (T - c)^{\frac{1}{p}}, k\right) \right\} \\
    \mathcal{I} &= \operatorname{unique\_int}\left(\left\{i_k \mid i_k=\left(l_k\right)^p+c, \text { where } l_k \in \mathcal{L}\right\} \right) \nonumber
\end{align}
where $\operatorname{linear\_space}(s, e, n)$ evenly spaces $n$ numbers from $s$ to $e$ (exclusive) and $\operatorname{unique\_int}(\cdot)$ convert the number to int and ensure the uniqueness of timesteps in the sequence. $c$ is the hyper-parameter for the selected center timestep. In this equation, the frequency of indexes changes in a quadratic manner as it moves away from a central timestep. It is essential to note that the aforementioned strategy represents one among several potential strategies. Alternative sequences, particularly centered on a specific timestep,  can also yield similar improvements in image quality.

%% file: sec/experiment.tex
\section{Experiment}

\subsection{Experimental Settings}

\paragraph{Models, Datasets and Evaluation Metrics}
To demonstrate the effectiveness of our method is agnostic with the type of pre-trained DMs, we evaluate our methods on three commonly used DMs: DDPM~\cite{ho2020ddpm}, LDM~\cite{rombach2022ldm} and Stable Diffusion~\cite{rombach2022ldm}\footnote{https://huggingface.co/runwayml/stable-diffusion-v1-5}.  Except for this, to show that our method is compatible with the fast sampling methods, we build our method upon 100-step DDIM~\cite{song2020ddim} for DDPM, 250-step for LDM and 50-step PLMS~\cite{liu2022pseudo} for Stable Diffusion, instead of the complete 1000-step denoising process. We select six datasets that are commonly adopted to evaluate these models, including CIFAR10~\cite{krizhevsky2009cifar10}, LSUN-Bedroom~\cite{yu2015lsun}, LSUN-Churches~\cite{yu2015lsun}, ImageNet~\cite{deng2009imagenet}, MS-COCO 2017~\cite{lin2014mscoco} and PartiPrompts~\cite{yu2022partiprompt}. For MS-COCO 2017 and PartiPrompt, we utilized the 5k validation set and 1.63k captions respectively as prompts for Stable Diffusion. For other datasets, we generate 50k images to assess the generation quality. We follow previous works~\cite{fang2023structural,yang2023diffusion, shih2023parallel} to employ the evaluation metrics including FID, sFID, IS, Precision-and-Recall and CLIP Score (on ViT-g/14)~\cite{heusel2017fid,hessel2021clipscore,kynkaanniemi2019precisionrecall,salimans2016IS}.




\begin{figure}[t]
    \centering
    \includegraphics[width=0.9\linewidth]{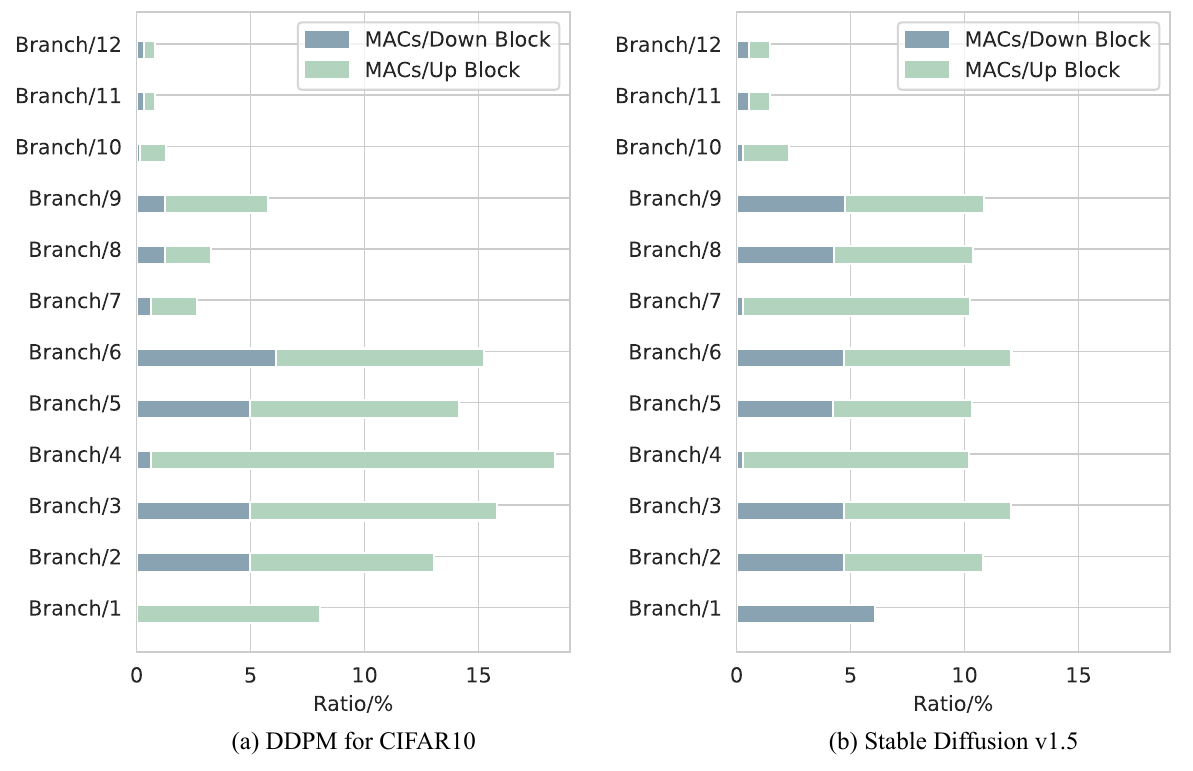}
    \caption{MACs for each skip branch, evaluated on DDPM for CIFAR10 and Stable Diffusion V1.5.}
    \label{fig:complexity}
\end{figure}

\paragraph{Baselines} 
We choose Diff-Pruning~\cite{fang2023structural} as the main baseline for our method since Diff-Pruning also reduces the training effort for compressing DMs. For the experiment on LDM, we extend~\cite{yang2023diffusion} as another baseline to represent one of the best methods to re-design a lightweight diffusion model. For the experiment on Stable Diffusion, we choose BK-SDMs~\cite{kim2023bksdm}, which are trained on 2.3M LAION image-text pairs, as baselines of architecturally compression and distillation for Stable Diffusion.

\begin{table*}[t]
\centering
    \small
    \resizebox{\linewidth}{!}{
    \begin{tabular}{l | c c c c | c c c c c }
      \toprule
      \multicolumn{10}{c}{\bf ImageNet 256 $\times$ 256}  \\
      \bf Method & \bf MACs $\downarrow$ & \bf Throughput $\uparrow$ & \bf Speed $\uparrow$ & \bf Retrain & \bf FID $\downarrow$ & \bf sFID $\downarrow$ & \bf IS $\uparrow$& \bf Precision $\uparrow$& \bf Recall $\uparrow$  \\
      \midrule
      IDDPM~\cite{nichol2021iDDPM} & 1416.3G & - & - & \xmark &  12.26 & 5.42 & - & 70.0 & 62.0 \\
      ADM-G~\cite{dhariwal2021adm} & 1186.4G & - & - & \xmark &   4.59 & 5.25 & 186.70 & 82.0 & 52.0 \\
      
      LDM-4~\cite{rombach2022ldm}  & 99.82G & 0.178 & 1$\times$ & \xmark & 3.60 & - & 247.67 & 87.0 & 48.0  \\
      LDM-4* & 99.82G & 0.178 & 1$\times$ & \xmark & 3.37 & 5.14 & 204.56 & 82.71 & 53.86 \\
      \midrule
      Spectral DPM~\cite{yang2023diffusion} & 9.9G & - & - & \cmark  & 10.60 & - & - & - & - \\
      Diff-Pruning~\cite{fang2023structural}* & 52.71G &  0.269 &  1.51$\times$ & \cmark & 9.27{\tiny(9.16)} & 10.59 & 214.42{\tiny (201.81)} & 87.87 & 30.87 \\  
      \midrule
      \bf Uniform - N=2 & 52.12G & 0.334 & 1.88$\times$ & \xmark & 3.39 & 5.11 & 204.09 & 82.75 & 54.07  \\
      \bf Uniform - N=3 & 36.48G & 0.471 & 2.65$\times$ & \xmark & 3.44 & 5.11 & 202.79 & 82.65 & 53.81  \\
      \bf Uniform - N=5 & 23.50G & 0.733 & 4.12$\times$ & \xmark & 3.59 & 5.16 & 200.45 & 82.36 & 53.31 \\
      \bf Uniform - N=10 & 13.97G & 1.239 & 6.96$\times$ & \xmark & 4.41	& 5.57 & 191.11 & 81.26	& 51.53 \\
      \bf Uniform - N=20 & 9.39G & 1.876 & 10.54$\times$  & \xmark & 8.23	& 8.08 & 161.83	& 75.31	& 50.57 \\
      \midrule
      \bf NonUniform - N=10  & 13.97G & 1.239 & 6.96$\times$ & \xmark & 4.27 & 5.42 & 193.11 & 81.75 & 51.84 \\ 
      \bf NonUniform - N=20  & 9.39G & 1.876 & 10.54$\times$ & \xmark & 7.11 & 7.34 & 167.85 & 77.44 & 50.08 \\
      
      \bottomrule
    \end{tabular}
    }
    \caption{Class-conditional generation quality on ImageNet using LDM-4-G. The baselines here, as well as our methods, employ 250 DDIM steps. *We reproduce Diff-Pruning to have a comprehensive comparison and the official results are shown in brackets. }
    \label{tbl:main_ldm_imagenet}
    \vspace{-2mm}
\end{table*}

\begin{table}[t]
\centering
    \small
    \resizebox{\linewidth}{!}{
    \begin{tabular}{l | c c c c | c}
      \toprule
      \multicolumn{6}{c}{\bf CIFAR-10 32 $\times$ 32 } \\
      \midrule
      \bf Method & \bf MACs $\downarrow$ & \bf Throughput $\uparrow$ & \bf Speed $\uparrow$ & \bf Steps $\downarrow$ & \bf FID $\downarrow$ \\
      \midrule
      DDPM         & 6.1G & 9.79 & 1$\times$ & - & 4.19\\
      DDPM*  & 6.1G & 9.79 & 1$\times$ & - & 4.16  \\
      \midrule
      Diff-Pruning  & 3.4G & 13.45 & 1.37$\times$ & 100k & 5.29  \\
      \midrule
      \bf Ours - N=2  & 4.15G  & 13.73 & 1.40$\times$ & 0 & 4.35 \\
      \bf Ours - N=3  & 3.54G & 15.74 & 1.61$\times$ & 0 & 4.70 \\
      \bf Ours - N=5  & 3.01G & 18.11 & 1.85$\times$ & 0 & 5.73 \\
      \bf Ours - N=10 & 2.63G & 20.26 & 2.07$\times$ & 0 & 9.74 \\
      \midrule
      \multicolumn{6}{c}{\bf LSUN-Bedroom  256 $\times$ 256}\\
      \midrule
      \bf Method & \bf MACs $\downarrow$ & \bf Throughput $\uparrow$ & \bf Speed $\uparrow$ & \bf Steps $\downarrow$ & \bf FID $\downarrow$ \\
      \midrule
      DDPM           & 248.7G & 0.21 & 1$\times$ & - & 6.62 \\
      DDPM*    & 248.7G & 0.21 & 1$\times$ & - & 6.70  \\
      \midrule
      Diff-Pruning   & 138.8G & 0.31 & 1.48$\times$ & 200k & 18.60 \\
      \midrule
      \bf Ours - N=2  &  190.8G & 0.27 & 1.29$\times$ & 0 & 6.69 \\
      \bf Ours - N=3  &  172.3G & 0.30 & 1.43$\times$ & 0 & 7.20 \\
      \bf Ours - N=5  &  156.0G & 0.31 & 1.48$\times$ & 0 & 9.49 \\
      \midrule
      \multicolumn{6}{c}{\bf LSUN-Churches  256 $\times$ 256 } \\
      \midrule
      \bf Method & \bf MACs $\downarrow$ & \bf Throughput $\uparrow$ & \bf Speed $\uparrow$ & \bf Steps $\downarrow$ & \bf FID $\downarrow$ \\
      \midrule
      DDPM     & 248.7G & 0.21 & 1$\times$ & - & 10.58 \\
      DDPM*    & 248.7G & 0.21 & 1$\times$ & - & 10.87  \\
      \midrule
      Diff-Pruning   & 138.8G & 0.31 & 1.48$\times$ & 500k & 13.90 \\
      \midrule
      \bf Ours - N=2  &  190.8G & 0.27 & 1.29$\times$ & 0 & 11.31 \\
      \bf Ours - N=3  &  172.3G & 0.30 & 1.43$\times$ & 0 & 11.75 \\
      \bf Ours - N=5  &  156.0G & 0.31 & 1.48$\times$ & 0 & 13.68 \\
      \bottomrule
    \end{tabular}
    }
    \caption{Results on CIFAR-10, LSUN-Bedroom and LSUN-Churches. All the methods here adopt 100 DDIM steps. * means the reproduced results, which are more comparable with our results since the random seed is the same.}
    \vspace{-5mm}
    \label{tbl:main_cifar}
\end{table}

\subsection{Complexity Analysis}  

We first analyze the improvement in inference speed facilitated by \methodname. 
The notable acceleration in inference speed primarily arises from incomplete reasoning in denoising steps, with layer removal accomplished by partitioning the U-Net by the skip connection. In Figure \ref{fig:complexity}, we present the division of MACs on two models. For each skip branch $i$, the MACs here contain the MACs in down block $D_i$ and the up block $U_i$. There is a difference in the amount of computation allocated to different skip branches for different models. Stable diffusion demonstrates a comparatively uniform distribution across layers, whereas DDPM exhibits more computational burden concentrated within the first several layers. Our approach would benefit from U-Net structures that have a larger number of skip branches, facilitating finer divisions of models, and giving us more choices for trade-off the speed and quality. In our experiment, we choose the skip branch 3/1/2 for DDPMs, LDM-4-G and Stable Diffusion respectively. We provide the results of using different branches in Appendix.


To comprehensively evaluate the efficiency of our method, in the following experiments, we report the throughput of each model using a single RTX2080 GPU. Besides, we report MACs in those tables, which refer to the average MACs for all steps.

\begin{figure*}[t]
    \centering
    \includegraphics[width=\linewidth]{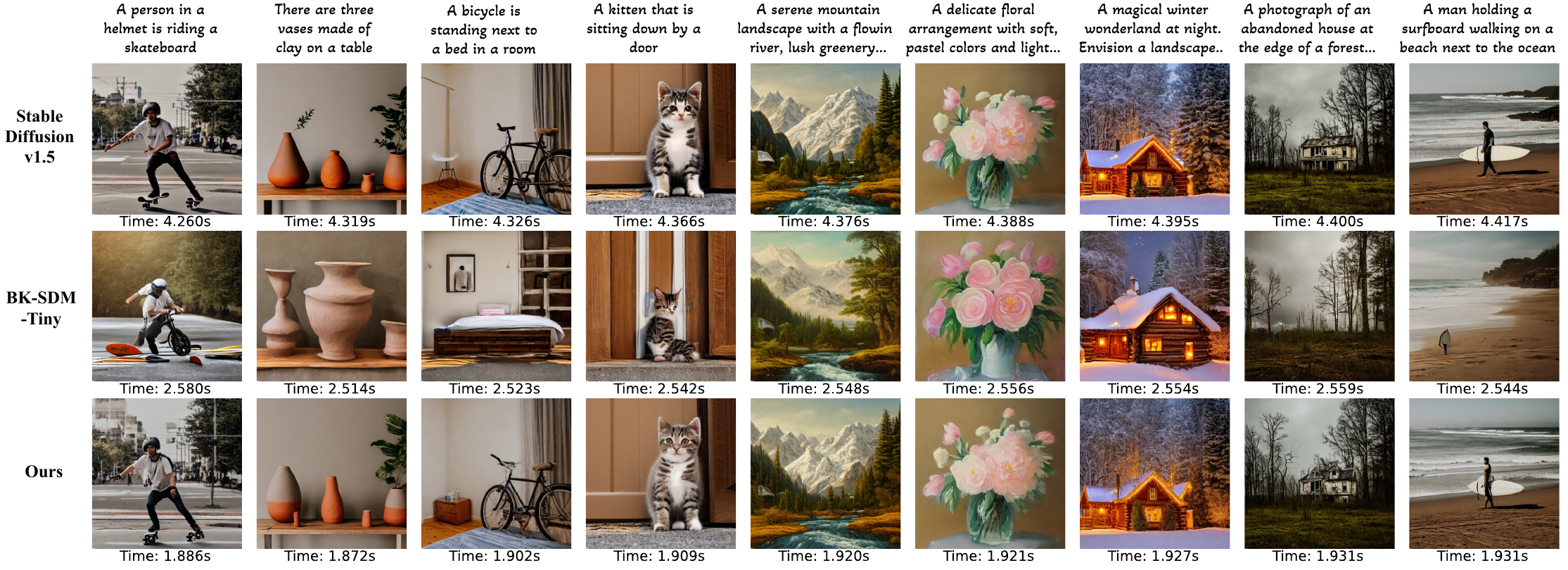}
    \caption{Visualization of the generated images by BK-SDM-Tiny and DeepCache. All the methods adopt the 50-step PLMS. The time here is the duration to generate a single image. Some prompts are omitted from this section for brevity. See Appendix for details.}
    \label{fig:examples}
\end{figure*}

\begin{table*}[h]
\centering
    \small
    \begin{tabular}{l | c c c c | c c c c}
      \toprule
      & \multicolumn{4}{c|}{\bf PartiPrompts} & \multicolumn{4}{c}{\bf COCO2017}\\
      \bf Method & \bf MACs $\downarrow$ & \bf Throughput $\uparrow$ & \bf Speed $\uparrow$ & \bf CLIP Score $\uparrow$ & \bf MACs $\downarrow$ & \bf Throughput $\uparrow$ & \bf Speed $\uparrow$ & \bf CLIP Score $\uparrow$ \\
      \midrule
      PLMS - 50 steps &  338.83G & 0.230 & 1.00$\times$ &  29.51 & 338.83G &  0.237 & 1.00$\times$ & 30.30\\
      \midrule
      PLMS - 25 steps &  169.42G & 0.470 & 2.04$\times$ & 29.33 & 169.42G & 0.453 & 1.91$\times$ &  29.99\\
      BK-SDM - Base &  223.81G & 0.343 & 1.49$\times$ & 28.88 &  223.81G & 0.344 & 1.45 $\times$ & 29.47 \\
      BK-SDM - Small & 217.78G & 0.402 & 1.75$\times$ & 27.94 & 217.78G & 0.397 & 1.68$\times$ & 27.96 \\
      BK-SDM - Tiny &  205.09G & 0.416 & 1.81$\times$ & 27.36 &  205.09G & 0.415 & 1.76 $\times$ & 27.41 \\
      \midrule
      \bf Ours            &  130.45G &  0.494 & 2.15$\times$ & 29.46 & 130.45G & 0.500 & 2.11$\times$ &  30.23 \\ 
      \bottomrule
    \end{tabular}
    \caption{Comparison with PLMS and BK-SDM. We utilized prompts in PartiPrompt and COCO2017 validation set to generate images at the resolution of 512. We choose N=5 to achieve a throughput that is comparable to or surpasses that of established baseline methods. Results for other choices of N can be found in Figure \ref{fig:parti_coco}. }
    \label{tbl:main_stable_diffusion}
\end{table*}

\subsection{Comparison with Compression Methods}

\paragraph{LDM-4-G for ImageNet.} We conduct experiments on ImageNet, and the results are shown in Table \ref{tbl:main_ldm_imagenet}. When accelerating to $4.1\times$ the speed, a minor performance decline is observed (from 3.39 to 3.59). Compared with the pruning and distillation methods, a notable improvement over those methods is observed in FID and sFID, even though the acceleration ratio of our method is more substantial. Furthermore, the augmentation in quality becomes more obvious with a larger number $N$ of caching intervals if we employ the non-uniform 1:N strategy. Detailed results for the non-uniform 1:N strategy for small N and the hyper-parameters for the non-uniform strategy are provided in the Appendix. 

\paragraph{DDPMs for CIFAR-10 and LSUN.} The results on CIFAR10, LSUN-Bedroom and LSUN-Churches are shown in Table \ref{tbl:main_cifar}. From these tables, we can find out that our method surpasses those requiring retraining, even though our methods have no retraining cost. Additionally, since we adopt a layer-pruning approach, which is more hardware-friendly, our acceleration ratio is more significant compared to the baseline method, under similar MACs constraints. 

\paragraph{Stable Diffusion.} The results are presented in Table \ref{tbl:main_stable_diffusion}. We outperform all three variants of BK-SDM, even with a faster denoising speed. As evident from the showcased examples in Figure \ref{fig:examples}, the images generated by our method exhibit a greater consistency with the images generated by the original diffusion model, and the image aligns better with the textual prompt.

\subsection{Comparison with Fast Sampler.}
\begin{figure}[h]
    \centering
    \includegraphics[width=0.47\linewidth]{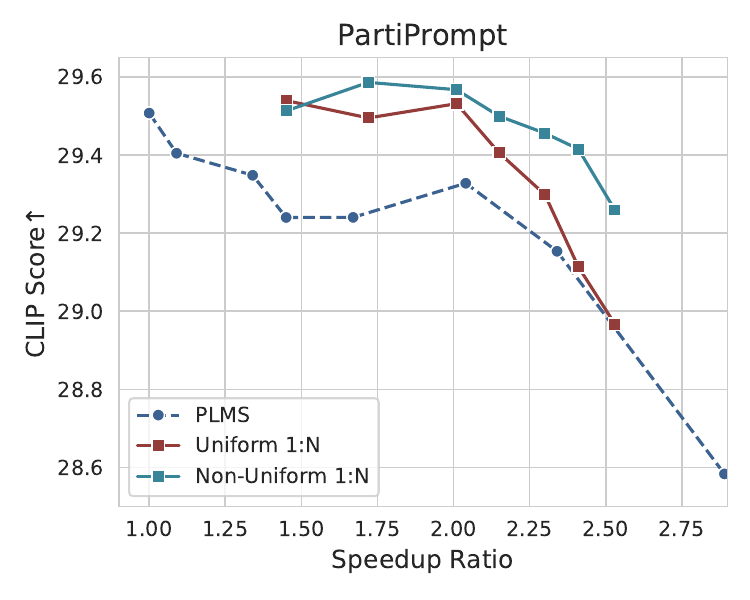}
    \includegraphics[width=0.47\linewidth]{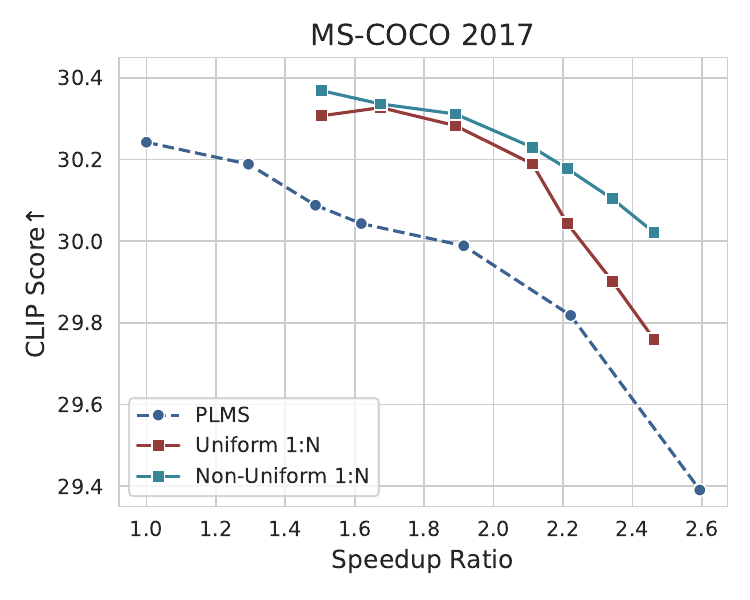}
    \caption{Comparison between PLMS, \methodname\ with uniform 1:N and non-uniform 1:N stratigies.}
    \label{fig:parti_coco}
\end{figure}

We conducted a comparative analysis with methods focused on reducing sampling steps. It is essential to highlight that our approach is additive to those fast samplers, as we show in previous experiments. 
In Table \ref{tbl:main_stable_diffusion} and Table \ref{tbl:compare_ddim}, we first compared our method with the DDIM~\cite{song2020ddim} or PLMS~\cite{liu2022pseudo} under similar throughputs. We observe that our method achieved slightly better results than 25-step PLMS on Stable Diffusion and comparable results to DDIM on LDM-4-G. We also measured the performance comparison between PLMS and our method under different acceleration ratios in Figure \ref{fig:parti_coco} to provide a more comprehensive comparison.

\begin{table}[h]
\centering
    \small
    \begin{tabular}{l | c | c c}
      \toprule
      \bf Method & \bf Throughput $\uparrow$ & \bf FID $\downarrow$ & \bf sFID $\downarrow$  \\
      \midrule
      DDIM - 59 steps & 0.727 & \bf 3.59 & \bf 5.14   \\
      \bf Ours            & 0.733 & \bf 3.59 & 5.16   \\ 
      \midrule
      DDIM - 91 steps & 0.436 & 3.46 & \bf 50.6   \\
      \bf Ours            & 0.471 & \bf 3.44 & 5.11   \\ 
      \bottomrule
    \end{tabular}
    \caption{Comparison with DDIM under the same throughput. Here we conduct class-conditional for ImageNet using LDM-4-G.}
    \label{tbl:compare_ddim}
\end{table}

\begin{figure*}[t]
    \centering
    \includegraphics[width=\linewidth]{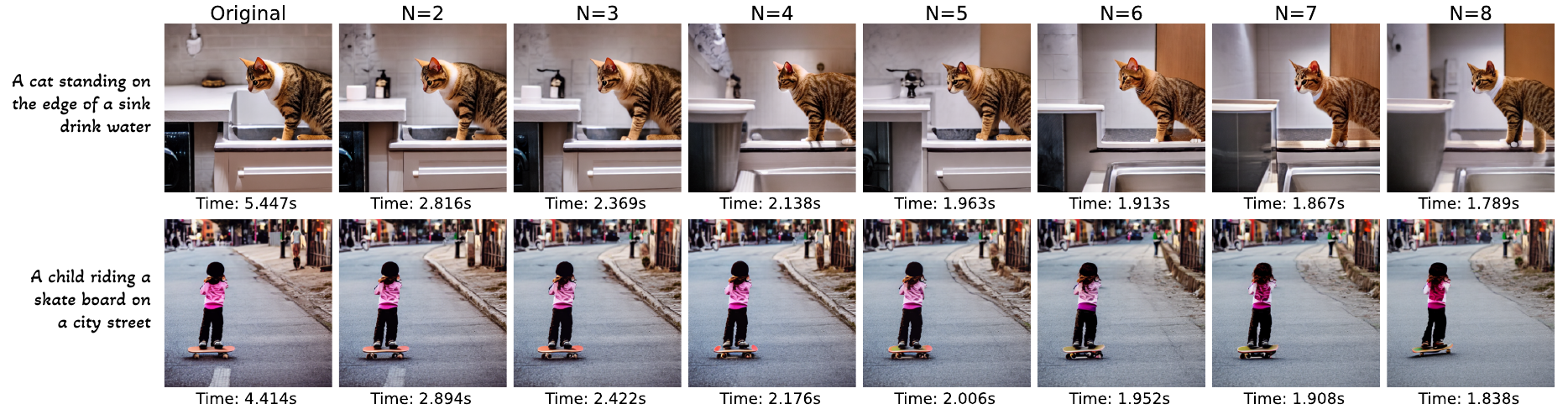}
    \caption{Illustration of the evolution in generated images with increasing caching interval N. }
    \label{fig:1_n}
\end{figure*}

\subsection{Analysis}

\paragraph{Ablation Study.}
DeepCache can be conceptualized as incorporating $(N-1) \times K$ steps of shallow network inference on top of the DDIM's K steps, along with more updates of the noisy images. It is important to validate whether the additional computations of shallow network inference and the caching of features yield positive effectiveness:
1) \textbf{Effectiveness of Cached Features:} We assess the impact of cached features in Table \ref{tbl:ablation_reuse}. Remarkably, we observe that, without any retraining, the cached features play a pivotal role in the effective denoising of diffusion models employing a shallow U-Net. 2) \textbf{Positive Impact of Shallow Network Inference:} Building upon the cached features, the shallow network inference we conduct has a positive impact compared to DDIM. Results presented in Table \ref{tbl:ablation_shallow} indicate that, with the additional computation of the shallow U-Net, DeepCache improves the 50-step DDIM by 0.32 and the 10-step DDIM by 2.98.


\paragraph{Illustration of the increasing caching interval N.} In Figure \ref{fig:1_n}, we illustrate the evolution of generated images as we increment the caching interval. A discernible trend emerges as a gradual reduction in time, revealing that the primary features of the images remain consistent with their predecessors. However, subtle details such as the color of clothing and the shape of the cat undergo modifications. Quantitative insights are provided in Table \ref{tbl:main_ldm_imagenet} and Figure \ref{fig:parti_coco}, where with an interval $N<5$, there is only a slight reduction in the quality of the generated image. 

\section{Limitations}
The primary limitation of our method originates from its dependence on the pre-defined structure of the pre-trained diffusion model. Specifically, when a model's shallowest skip branch encompasses a substantial portion of computation, such as 50\% of the whole model, the achievable speedup ratio through our approach becomes relatively constrained. 
Additionally, our method encounters non-negligible performance degradation with larger caching steps (e.g., N=20), which could impose constraints on the upper limit of the acceleration ratio.

\begin{table}[t]
\centering
    \small
    \resizebox{\linewidth}{!}{
    \begin{tabular}{l l | c c }
      \toprule
      \bf Model & \bf Dataset & \bf DeepCache & \bf w/o Cached Features   \\
      \midrule
      DDPM & Cifar10 & 9.74 & 192.98 \\
      LDM-4-G & ImageNet & 7.36 & 312.12 \\
      \bottomrule
    \end{tabular}
    }
    \caption{Effectiveness of Cached Features. Under identical hyperparameters, we replace the cached features with a zero matrix.}
    \label{tbl:ablation_reuse}
\end{table}

\begin{table}[t]
\centering
    \small
    \begin{tabular}{l | c c | c}
      \toprule
      \bf Steps & \bf DDIM FID$\downarrow$ & \bf \methodname\ FID$\downarrow$  & $\Delta$ \\
      \midrule
      50 & 4.67 & 4.35 & -0.32\\
      20 & 6.84 & 5.73 & -1.11\\
      10 & 13.36 & 10.38 & -2.98\\
      \bottomrule
    \end{tabular}
    \caption{Effectiveness of Shallow Network Inference. Steps here mean the number of steps that perform full model inference.}
    \label{tbl:ablation_shallow}
    \vspace{-3mm}
\end{table}

\section{Conclusion}
In this paper, we introduce a novel paradigm, DeepCache,
to accelerate the diffusion model. Our strategy employs the similarity observed in high-level features across adjacent steps of the diffusion model, thereby mitigating the computational overhead associated with redundant high-level feature calculations. Additionally, we leverage the structural attributes in the U-Net architecture to facilitate the updating of low-level features. Through the adoption of DeepCache, a noteworthy acceleration in computational speed is achieved. Empirical evaluations on several datasets and diffusion models demonstrate that DeepCache surpass other compression methods that focus on the reduction of parameter size. Moreover, the proposed algorithm demonstrates comparable and even slightly superior generation quality compared to existing techniques such as DDIM and PLMS, thereby offering a new perspective in the field.

%% file: sec/X_suppl.tex
\clearpage
\setcounter{section}{0}
\maketitlesupplementary

\renewcommand{\thesection}{\Alph{section}}
\section{Pseudo algorithm}

We present the pseudocode for our algorithm in Algorithm \ref{alg:main}. It illustrates the iterative generation process over N steps, involving one step of complete model inference and N-1 steps of partial model inference. Here, we employ the sampling algorithm from DDPM~\cite{ho2020ddpm} as an example. Our algorithm is adaptable to other fast sampling methods.

\algdef{SE}[SUBALG]{Indent}{EndIndent}{}{\algorithmicend\ }%
\algtext*{Indent}
\algtext*{EndIndent}
\definecolor{myblue}{rgb}{0.7,0.7,0.7}
\algnewcommand{\MyLineComment}[1]{\textcolor{gray}{\(\triangleright\) #1}}

\begin{algorithm}[h]
    \small
    \LinesNumberedHidden

   \KwIn{A U-Net Model with down-sample blocks $\{D_i\}_{i=1}^d$, up-sample blocks$\{U_i\}_{i=1}^d$ and middle blocks $M$}
   \KwIn{Caching Interval $N$, Branch Index $m$}
   \KwIn{Output from step $x_t$, timestep $t$}
   
   \KwOut{predicted output at $t-N$ step}

    \MyLineComment{1. Cache Step -  Calculate $\boldsymbol{\epsilon}_\theta\left(\mathbf{x}_t, t\right)$ and $x_{t-1}$}

    $\mathbf{h}_0 \leftarrow \mathbf{x}_t$ \Comment{{\scriptsize $\mathbf{h}_{i}$ for down-sampling features}}
    
    \For {$i = 1,\ldots,d$}
    {
        $\mathbf{h}_i \leftarrow D_i(\mathbf{h}_{i-1})$
    }
    
    $\mathbf{u}_{d+1} \leftarrow M(\mathbf{h}_{d})$ \Comment{{\scriptsize $\mathbf{u}_{i}$ for up-sampling features}}

    \For {$i = d,\ldots,1$}
    {
        \If {$i = m$} 
        {
            Store $\mathbf{u}_{i+1}$ in Cache 
        }

        $\mathbf{u}_i \leftarrow U_i(\text{Concat}(\mathbf{u}_{i+1}, \mathbf{h}_{i}))$
    }
    
    
    $\mathbf{x}_{t-1}=\frac{1}{\sqrt{\alpha_t}}\left(\mathbf{x}_t-\frac{1-\alpha_t}{\sqrt{1-\bar{\alpha}_t}} 
    \mathbf{u}_1
    \right)+\sigma_t \mathbf{z}$ \Comment{{\scriptsize $\mathbf{z} \sim \mathcal{N}(\mathbf{0}, \mathbf{I})$}}

    \MyLineComment{2. Retrieve Step - Calculate $\{x_{t-i}\}_{i=2}^N$}

    \For {$n = 2,\ldots,N$}
    {
        $\mathbf{h}_0 \leftarrow \mathbf{x}_{t-n+1}$ 
    
        \For {$i = 1,\ldots,m$}
        {
            $\mathbf{h}_i \leftarrow D_i(\mathbf{h}_{i-1})$
        }
    
        Retrieve $\mathbf{u}_{i+1}$ from Cache
        
        \For {$i = m,\ldots,1$}
        {  
            $\mathbf{u}_i \leftarrow U_i(\text{Concat}(\mathbf{u}_{i+1}, \mathbf{h}_{i}))$
        }
        
        
        $\mathbf{x}_{t-n}=\frac{1}{\sqrt{\alpha_t}}\left(\mathbf{x}_t-\frac{1-\alpha_t}{\sqrt{1-\bar{\alpha}_t}} 
        \mathbf{u}_1
        \right)+\sigma_t \mathbf{z}$  \Comment{{\scriptsize $\mathbf{z} \sim \mathcal{N}(\mathbf{0}, \mathbf{I})$}}
    }
    return $x_{t-N}$ 
    
   \caption{DeepCache}\label{alg:main}
\end{algorithm}

\section{Varying Hyper-parameters in Non-uniform 1:N Strategy} 

In the non-uniform 1:N strategy, the two hyper-parameters involved are the center $c$ and the power $p$, which is used to determine the sequence of timesteps for conducting the entire model inference. We test on LDM-4-G for the impact of these hyper-parameters. Results are shown in Table \ref{tbl:varying_c} and Table \ref{tbl:varying_p}. From these two tables, a clear trend is evident in the observations: as the parameters $p$ and $c$ are incremented, there is an initial improvement in the generated image quality followed by a subsequent decline. This pattern affirms the effectiveness of the strategy and also aligns with the location of the significant decrease in similarity observed in Figure \ref{fig:similarity}(c).

\begin{table}[htbp!]
\centering
    \small
    \resizebox{\linewidth}{!}{
    \begin{tabular}{l | c c c c c }
      \toprule
      \multicolumn{6}{c}{\bf ImageNet 256 $\times$ 256}  \\
      \bf Center & \bf FID $\downarrow$ & \bf sFID $\downarrow$ & \bf IS $\uparrow$& \bf Precision $\uparrow$& \bf Recall $\uparrow$  \\ 
      \midrule
        $c$ = 10	& 8.26	& 8.47	& 160.3	    & 75.69	& 48.93 \\
        $c$ = 20	& 8.17	& 8.46	& 161.18	& 75.77	& 48.95 \\
        $c$ = 50	& 7.77	& 8.16	& 163.74	& 76.23	& 49.18 \\
        $c$ = 80	& 7.36	& 7.76	& 166.21	& 76.93	& 49.75 \\
        $c$ = 100	& 7.16	& 7.51	& 167.52	& 77.30	& 49.64 \\
        $c$ = 120	& \bf 7.11	& \bf 7.34	& \bf 167.85	& \bf 77.44	& \bf 50.08 \\
        $c$ = 150	& 7.33	& 7.36	& 166.04	& 77.09	& 49.98 \\
        $c$ = 200	& 8.09  & 7.79  & 160.50    & 75.85 & 49.69 \\
      \bottomrule
    \end{tabular}
    }
    \caption{Varying Center $c$ with the power $p$ equals to 1.2. Here the caching interval is set to 20.}
    \label{tbl:varying_c}
    \vspace{-2mm}
\end{table}

\begin{table}[h!]
\centering
    \small
    \resizebox{\linewidth}{!}{
    \begin{tabular}{l | c c c c c }
      \toprule
      \multicolumn{6}{c}{\bf ImageNet 256 $\times$ 256}  \\
      \bf Power & \bf FID $\downarrow$ & \bf sFID $\downarrow$ & \bf IS $\uparrow$& \bf Precision $\uparrow$& \bf Recall $\uparrow$  \\ 
      \midrule
        $p$ = 1.05	&  7.36	& 7.52	& 166.12	& 77.06	& 50.38\\
        $p$ = 1.1	&  7.25 & 7.44  & 166.82    & 77.17 & 50.13\\
        $p$ = 1.2	&  7.11	& \bf 7.34	& 167.85	& 77.44	& 50.08\\
        $p$ = 1.3	&  \bf 7.09	& 7.35	& \bf 167.97	& \bf 77.56	& \bf 50.34\\
        $p$ = 1.4	&  7.13	& 7.39	& 167.68	& 77.42	& 50.26\\
        $p$ = 1.5	&  7.25 & 7.44  & 166.82    & 77.17 & 50.13  \\
      \bottomrule
    \end{tabular}
    }
    \caption{Varying Power $p$ with the center $c$ equals to 120. Here the caching interval is also set to 20. }
    \vspace{-2mm}
    \label{tbl:varying_p}
\end{table}

\begin{table}[h]
\centering
\small
    \resizebox{0.8\linewidth}{!}{
    \begin{tabular}{l | c c c c c }
      \toprule
      & N=2 & N=3 & N=5 & N=10 & N=20 \\
      \midrule
      Center - $c$ & 120  & 120 & 110 & 110 & 120 \\
      Power - $p$ &  1.2  & 1.2 & 1.4 & 1.2 & 1.4 \\
    \bottomrule
    \end{tabular}
    }
    \caption{Hyper-parameters for the non-uniform 1:N strategy in LDM-4-G}
    \label{tbl:hyper_center_pow_ldm}
    \vspace{-4mm}
\end{table}
 
\begin{table}[h]
\centering
    \resizebox{0.8\linewidth}{!}{
    \begin{tabular}{l | c c c c c c c c}
      \toprule
      & N=2 & N=3 & N=4 & N=5 & N=6 & N=7 & N=8 \\
      \midrule
      Center - $c$ & 15 & 15 & 15 & 10 & 15 & 15 & 10 \\
      Power - $p$ & 1.5 & 1.3 & 1.4 & 1.5 & 1.3 & 1.4 & 1.4  \\
      \midrule
      Center - $c$ & 20 & 20 & 20 & 15 & 15 & 15 & 20 \\
      Power - $p$ & 1.3 & 1.4 & 1.4 & 1.3 & 1.5 & 1.5 & 1.3 \\  
    \bottomrule
    \end{tabular}
    }
    \caption{Hyper-parameters for the non-uniform 1:N strategy in Stable Diffusion v1.5.}
    \label{tbl:hyper_center_pow}
    \vspace{-4mm}
\end{table}

\begin{table*}[t]
\centering
    \small
    \resizebox{\linewidth}{!}{
    \begin{tabular}{l | c c c c c | l | c c c c c}
      \toprule
      \multicolumn{12}{c}{\bf ImageNet 256 $\times$ 256 (250 DDIM Steps)}  \\
      \bf Method & \bf FID $\downarrow$ & \bf sFID $\downarrow$ & \bf IS $\uparrow$& \bf Precision $\uparrow$& \bf Recall $\uparrow$ & \bf Method & \bf FID $\downarrow$ & \bf sFID $\downarrow$ & \bf IS $\uparrow$& \bf Precision $\uparrow$& \bf Recall $\uparrow$ \\
      \midrule
      Baseline - LDM-4* & 3.37 & 5.14 & 204.56 & 82.71 & 53.86 & Baseline - LDM-4 & 3.60 & - & 247.67 & 87.0 & 48.0 \\
      \midrule
      \bf Uniform - N=2 & 3.39 & 5.11 & 204.09 & 82.75 & 54.07  & Non-uniform - N=2  & 3.46 & 5.14 & 204.12 & 83.21 & 53.53 \\
      \bf Uniform - N=3 & 3.44 & 5.11 & 202.79 & 82.65 & 53.81  & Non-uniform - N=3  & 3.49 & 5.13 & 203.22 & 83.18 & 53.44 \\
      \bf Uniform - N=5 & 3.59 & 5.16 & 200.45 & 82.36 & 53.31 & Non-uniform - N=5  & 3.63 & 5.12 & 200.04 & 83.07 & 53.25 \\
      \bf Uniform - N=10 & 4.41	& 5.57 & 191.11 & 81.26	& 51.53 & Non-uniform - N=10  & 4.27 & 5.42 & 193.11 & 81.75 & 51.84 \\
      \bf Uniform - N=20 & 8.23	& 8.08 & 161.83	& 75.31	& 50.57 & Non-uniform - N=20 & 7.36 & 7.76 & 166.21 & 76.93 & 49.75 \\
      \bottomrule
    \end{tabular}
    }
    \caption{Comparing non-uniform and uniform 1:N strategy in class-conditional generation for ImageNet using LDM-4-G. *We regenerate the images using the official checkpoint of LDM-4-G.}
    \label{tbl:imagenet_uniform_non_uniform}
    \vspace{-2mm}
\end{table*}

\paragraph{Selected Hyper-parameters for non-uniform 1:N}

For experiments in LDM, the optimal hyper-parameters and shown in Table \ref{tbl:hyper_center_pow_ldm}. 
For experiments in Stable Diffusion, we chose center timesteps from the set \{0, 5, 10, 15, 20, 25\} and power values from the set \{1.1, 1.2, 1.3, 1.4, 1.5, 1.6\}. The optimal hyper-parameter values employed in our experiments are detailed in Table \ref{tbl:hyper_center_pow}. 

From the selected hyper-parameters, we found out that the optimal values vary slightly across different datasets. A noticeable trend is observed, indicating that the majority of optimal parameters tend to center around the 15th timestep, accompanied by a power value of approximately 1.4.

\section{Non-uniform 1:N v.s. Uniform 1:N}

We have shown the comparison of the non-uniform 1:N versus uniform 1:N strategy on Stable Diffusion in Figure \ref{fig:parti_coco}. Here, we extend the comparison to ImageNet with LDM-4-G, and the corresponding results are detailed in Table \ref{tbl:imagenet_uniform_non_uniform}.

In accordance with the observations from Table \ref{tbl:imagenet_uniform_non_uniform}, a consistent pattern emerges compared to the findings on Stable Diffusion. Notably, when employing a substantial caching interval, the non-uniform strategy demonstrates a notable improvement, with the FID increasing from 8.23 to 7.36 with N=20. However, when dealing with a smaller caching interval (N$<$5), the strategy does not yield an enhancement in image quality. In fact, in certain cases, it may even lead to a slight degradation of images, as evidenced by the FID increasing from 3.39 to 3.46 for N=2. 

\section{Varying Skip Branches}

In Table \ref{tbl:cifar_varying_branch}, we show the impact on image quality as we vary the skip branch for executing DeepCache. For our experiments, we employ the uniform 1:N strategy with N=5, and the sampling of DDIM still takes 100 steps. 
From the results in the table, we observe that the choice of skip branch introduces a trade-off between speed and image fidelity. Specifically, opting for the first skip branch with no down-sampling blocks and one up-sampling block yields approximately 3$\times$ acceleration, accompanied by a reduction in FID to 7.14. Additionally, certain skip branches exhibit significant performance variations, particularly the 6-th branch.
The results emphasize an extra trade-off between speed and image quality, complementing the earlier noted trade-off linked to different sampling steps. This particular trade-off operates at the level of model size granularity and can be achieved without incurring additional costs.

\begin{table}[t]
\centering
    \small
    \resizebox{\linewidth}{!}{
    \begin{tabular}{c | c c c | c}
      \toprule
      \multicolumn{5}{c}{\bf CIFAR-10 32 $\times$ 32} \\
      \midrule
      \bf Skip Branch& \bf MACs $\downarrow$ & \bf Throughput $\uparrow$ & \bf Speed $\uparrow$ &  \bf FID $\downarrow$ \\
      \midrule
      1 &  1.60G & 29.60 & 3.023$\times$ & 7.14\\
      2 &  2.24G & 22.24 & 2.272$\times$ & 5.94\\
      3 &  3.01G & 18.11 & 1.850$\times$ & 5.73\\
      4 &  3.89G & 15.44 & 1.577$\times$ & 5.69\\
      5 &  4.58G & 13.15 & 1.343$\times$ & 5.51\\
      6 &  5.31G & 11.46 & 1.171$\times$ & 4.93\\
      7 &  5.45G & 11.27 & 1.151$\times$ & 4.92\\
      8 &  5.60G & 11.07 & 1.131$\times$ & 4.76\\
      9 &  5.88G & 10.82 & 1.105$\times$ & 4.54\\
      10 & 5.95G & 10.73 & 1.096$\times$ & 4.57\\
      11 & 5.99G & 10.67 & 1.089$\times$ & 4.52\\
      12 & 6.03G & 10.59 & 1.082$\times$ & 4.48\\
      \bottomrule
    \end{tabular}
    }
    \caption{Effect of different skip branches. Here we test the impact under the uniform 1:5 strategy.}
    \vspace{-5mm}
    \label{tbl:cifar_varying_branch}
\end{table}

\begin{table*}[t]
\centering
    \resizebox{0.8\linewidth}{!}{
    \begin{tabular}{l | c c | c | l | c c | c c}
      \toprule
      \multicolumn{4}{c|}{\bf PLMS} & \multicolumn{5}{c}{\bf DeepCache} \\
       Steps & Throughput & Speed & CLIP Score & N & Throughput & Speed &  Uniform 1:N & Non-Uniform 1:N\\
      \midrule
      50 & 0.230 & 1.00 & 29.51 & 1 & - & - & - & - \\
      45 & 0.251 & 1.09 & 29.40 & 2 & 0.333 & 1.45 & 29.54 & 29.51 \\
      40 & 0.307 & 1.34 & 29.35 & 3 & 0.396 & 1.72 & 29.50 & 29.59 \\
      35 & 0.333 & 1.45 & 29.24 & 4 & 0.462 & 2.01 & 29.53 & 29.57 \\
      30 & 0.384 & 1.67 & 29.24 & 5 & 0.494 & 2.15 & 29.41 & 29.50 \\
      25 & 0.470 & 2.04 & 29.32 & 6 & 0.529 & 2.30 & 29.30 & 29.46 \\
      20 & 0.538 & 2.34 & 29.15 & 7 & 0.555 & 2.41 & 29.11 & 29.42 \\
      15 & 0.664 & 2.89 & 28.58 & 8 & 0.582 & 2.53 & 28.97 & 29.26 \\
    \bottomrule
    \end{tabular}
    }
    \caption{Stable Diffusion v1.5 on PartiPrompt}
    \label{tbl:parti}
\end{table*}

\begin{table*}[t]
\centering
\resizebox{0.8\linewidth}{!}{
    \begin{tabular}{l | c c | c | l | c c | c c}
      \toprule
      \multicolumn{4}{c|}{\bf PLMS} & \multicolumn{5}{c}{\bf DeepCache} \\
       Steps & Throughput & Speed & CLIP Score & N & Throughput & Speed &  Uniform 1:N & Non-Uniform 1:N\\
      \midrule
      50 & 0.237 & 1.00 & 30.24 & 1 & - & - & - & -  \\
      45 & 0.252 & 1.06 & 30.14 & 2 & 0.356 & 1.50 & 30.31 & 30.37 \\
      40 & 0.306 & 1.29 & 30.19 & 3 & 0.397 & 1.68 & 30.33 & 30.34 \\
      35 & 0.352 & 1.49 & 30.09 & 4 & 0.448 & 1.89 & 30.28 & 30.31 \\
      30 & 0.384 & 1.62 & 30.04 & 5 & 0.500 & 2.11 & 30.19 & 30.23 \\
      25 & 0.453 & 1.91 & 29.99 & 6 & 0.524 & 2.21 & 30.04 & 30.18 \\
      20 & 0.526 & 2.22 & 29.82 & 7 & 0.555 & 2.34 & 29.90 & 30.10 \\
      15 & 0.614 & 2.59 & 29.39 & 8 & 0.583 & 2.46 & 29.76 & 30.02 \\
    \bottomrule
    \end{tabular}
    }
    \caption{Stable Diffusion v1.5 on MS-COCO 2017}
    \label{tbl:coco}
\end{table*}

\begin{figure*}[t!]
    \centering
    \includegraphics[width=\linewidth]{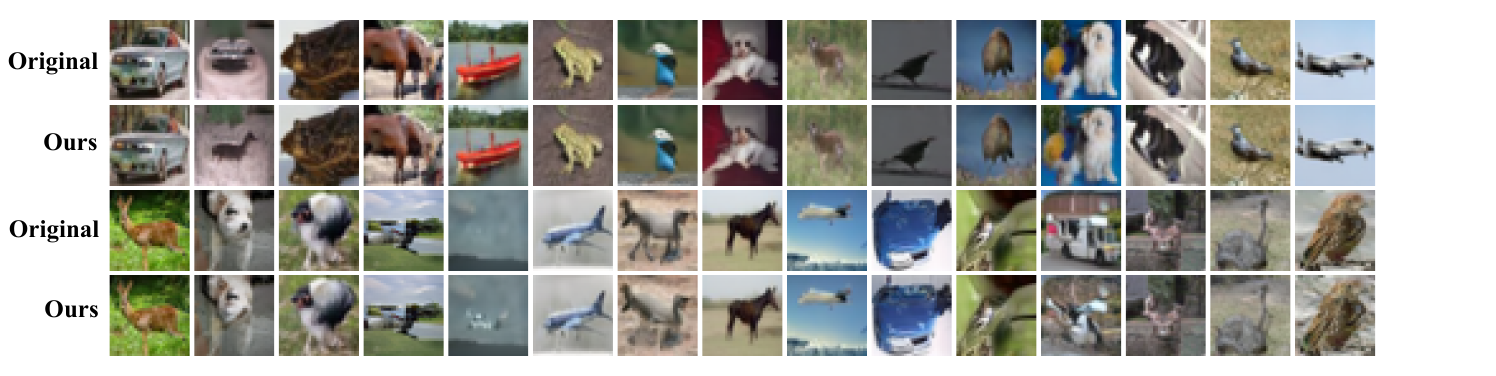}
    \caption{DDPM for LSUN-Churches: Samples with DDIM-100 steps (upper line) and DDIM-100 steps + DeepCache with N=5 (lower line). The speedup Ratio here is 1.85$\times$.}
    \label{fig:cifar_samples}
\end{figure*}

\section{Prompts}
Prompts in  Figure \ref{fig:teaser}(a):
\begin{itemize}
    \item A bustling city street under the shine of a full moon
    \item A picture of a snowy mountain peak, illuminated by the first light of dawn
    \item dark room with volumetric light god rays shining through window onto stone fireplace in front of cloth couch
    \item A photo of an astronaut on a moon
    \item A digital illustration of a medieval town, 4k, detailed, trending in artstation, fantasy
    \item A photo of a cat. Focus light and create sharp, defined edges
\end{itemize}

Prompts in Figure \ref{fig:examples}:
\begin{itemize}
    \item A person in a helmet is riding a skateboard
    \item There are three vases made of clay on a table
    \item A very thick pizza is on a plate with one piece taken.
    \item A bicycle is standing next to a bed in a room.
    \item A kitten that is sitting down by a  door.
    \item A serene mountain landscape with a flowing river, lush greenery, and a backdrop of snow-capped peaks, in the style of an oil painting.
    \item A delicate floral arrangement with soft, pastel colors and light, flowing brushstrokes typical of watercolor paintings.
    \item A magical winter wonderland at night. Envision a landscape covered in fresh snow, with twinkling stars above, a cozy cabin with smoke rising from its chimney, and a gentle glow from lanterns hung on the trees
    \item A photograph of an abandoned house at the edge of a forest, with lights mysteriously glowing from the windows, set against a backdrop of a stormy sky. high quality photography, Canon EOS R3.
    \item A man holding a surfboard walking on a beach next to the ocean.
\end{itemize}

\section{Detailed Results for Stable Diffusion}
We furnish the elaborate results corresponding to Figure \ref{fig:parti_coco} in Table \ref{tbl:parti} and Table \ref{tbl:coco}. Given the absence of a definitive N for aligning the throughput of PLMS, we opt for an alternative approach by exploring results for various N values. Additionally, we assess the performance of the PLMS algorithm across different steps. Analyzing the data from these tables reveals that for N $<$ 5, there is minimal variation in the content of the image, accompanied by only slight fluctuations in the CLIP Score.

\section{More Samples for Each Dataset}
\begin{figure*}[t]
    \centering
    \includegraphics[width=\linewidth]{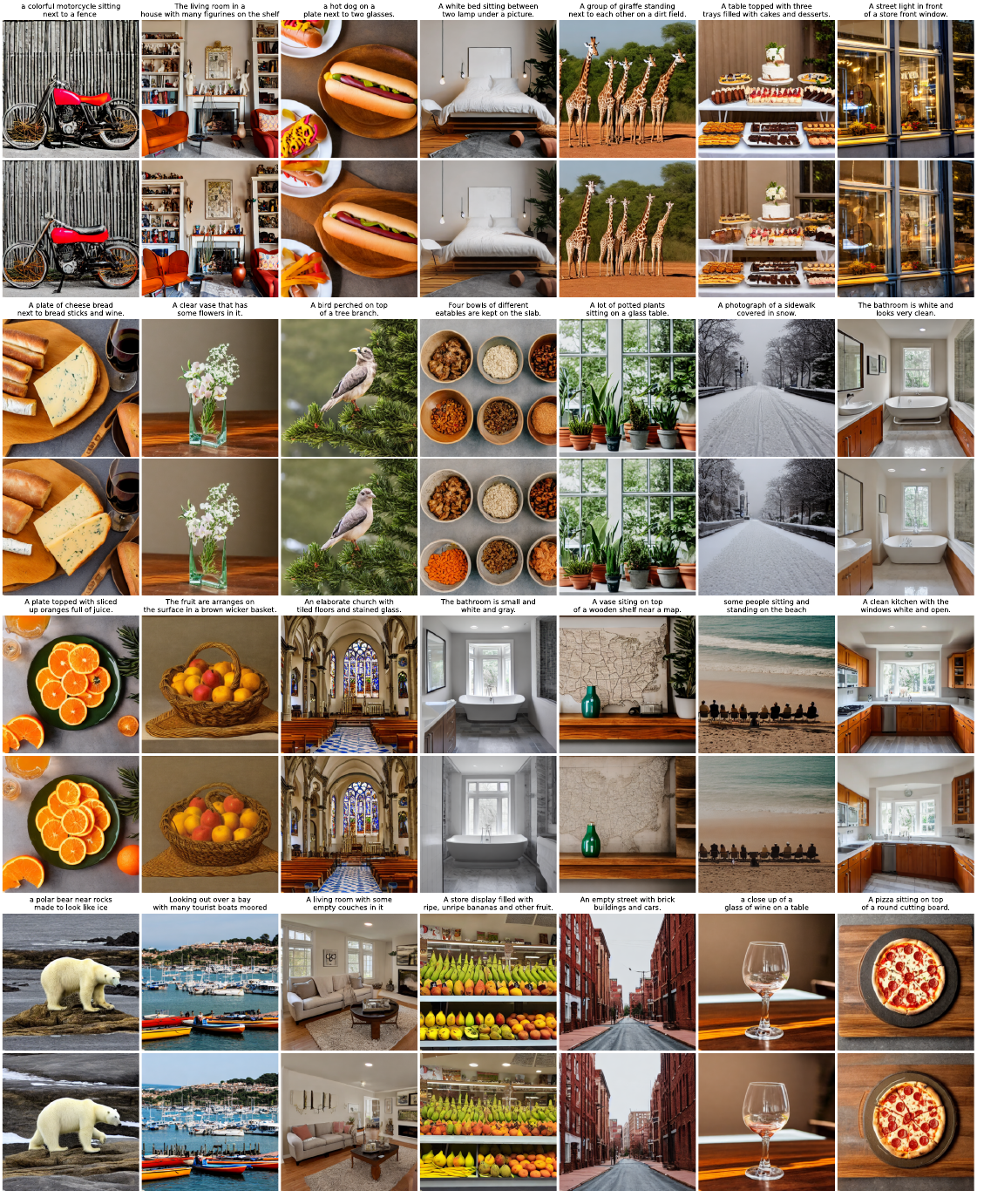}
    \caption{Stable Diffusion v1.5: Samples with 50 PLMS steps (upper line) and 50 PLMS steps + DeepCache with N=5 (lower line). The speedup Ratio here is 2.15$\times$. Here we select prompts from the MS-COCO 2017 validation set.}
    \label{fig:sdm_samples}
\end{figure*}

\begin{figure*}[t]
    \centering
    \includegraphics[width=\linewidth]{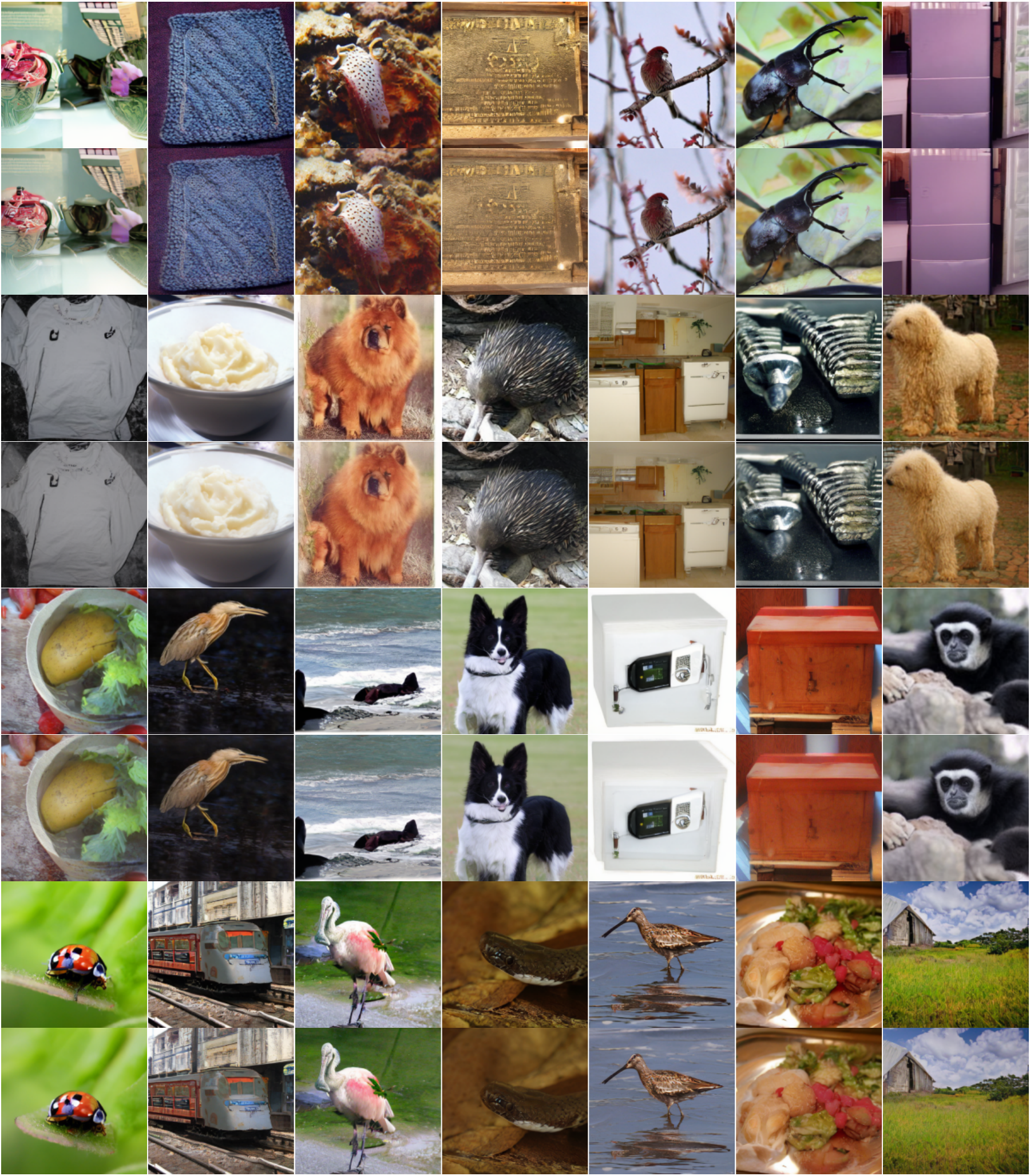}
    \caption{LDM-4-G for ImageNet: Samples with DDIM-250 steps (upper line) and DDIM-250 steps + DeepCache with N=10 (lower line). The speedup Ratio here is 6.96$\times$.}
    \label{fig:imagenet_samples}
\end{figure*}

\begin{figure*}[t]
    \centering
    \includegraphics[width=\linewidth]{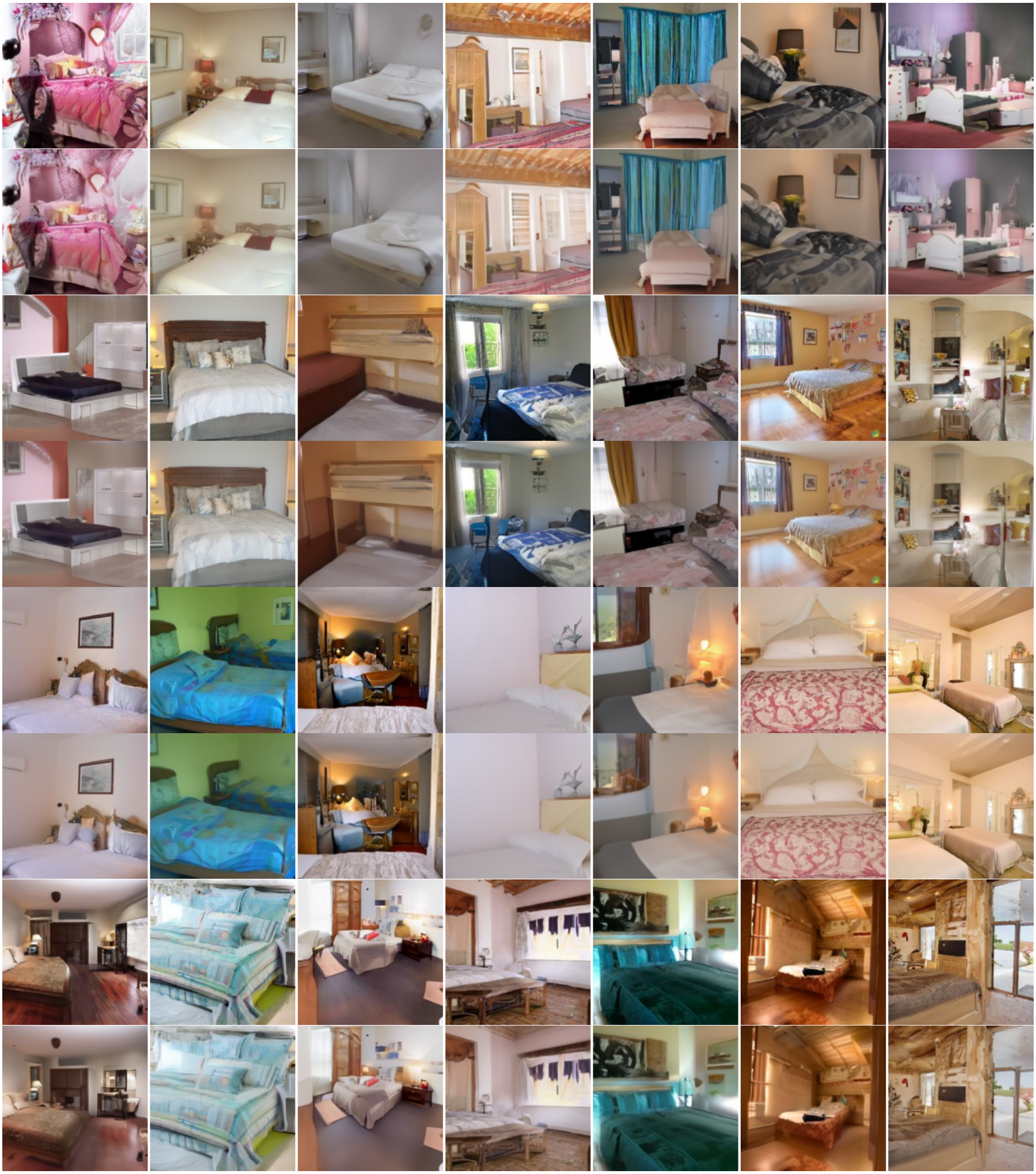}
    \caption{DDPM for LSUN-Bedroom: Samples with DDIM-100 steps (upper line) and DDIM-100 steps + DeepCache with N=5 (lower line). The speedup Ratio here is 1.48$\times$.}
    \label{fig:bedroom_samples}
\end{figure*}

\begin{figure*}[t]
    \centering
    \includegraphics[width=\linewidth]{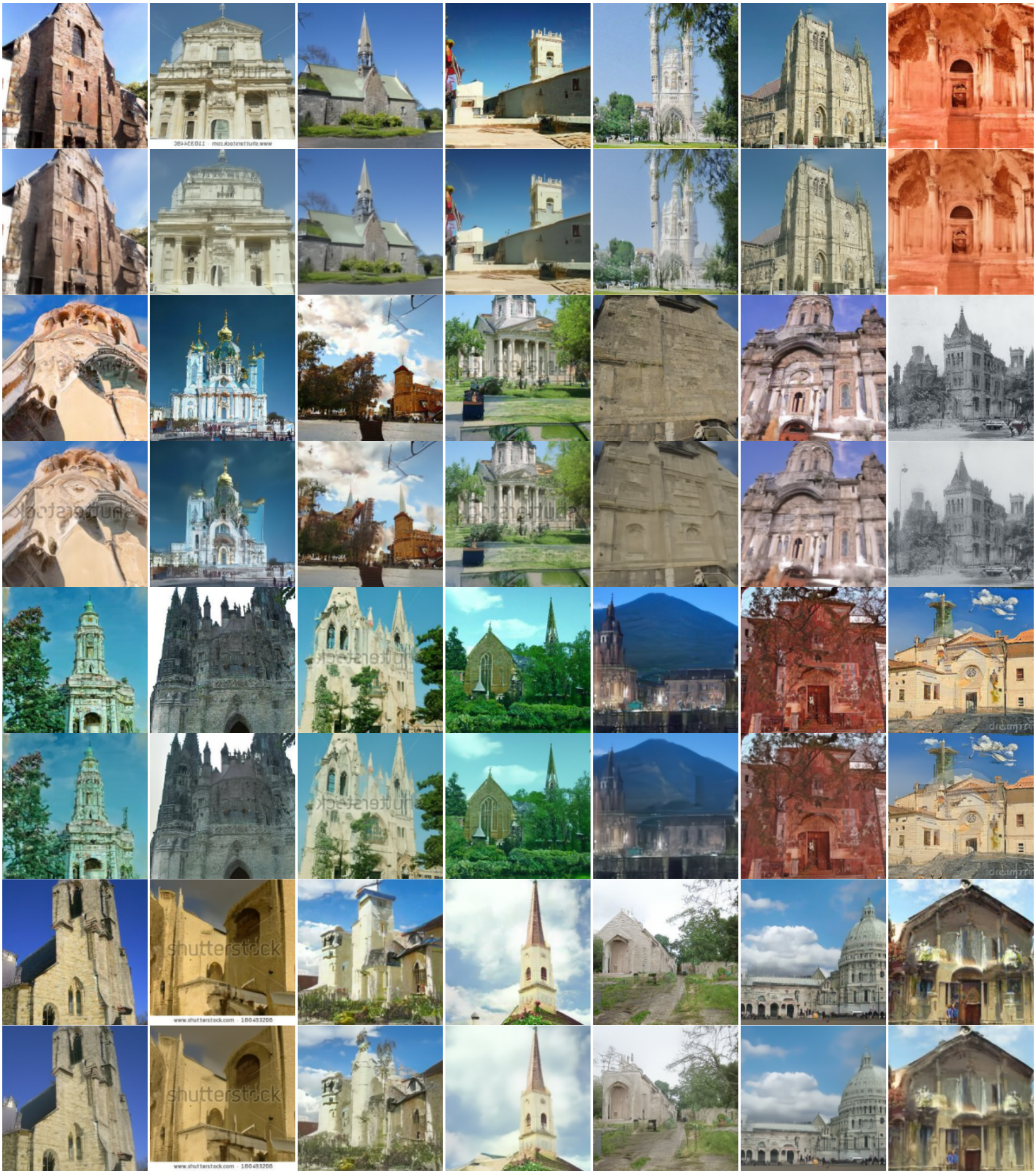}
    \caption{DDPM for LSUN-Churches: Samples with DDIM-100 steps (upper line) and DDIM-100 steps + DeepCache with N=5 (lower line). The speedup Ratio here is 1.48$\times$. }
    \label{fig:church_samples}
\end{figure*}

We provide the generated images for each model and each dataset in Figure \ref{fig:cifar_samples}, Figure \ref{fig:sdm_samples}, Figure \ref{fig:imagenet_samples}, Figure \ref{fig:bedroom_samples} and Figure \ref{fig:church_samples}.

